\documentclass[12pt]{article}
\usepackage{authblk}
\usepackage{natbib}
\usepackage{bm, amsmath,amssymb,amsfonts}
\usepackage{hyperref}
\usepackage{algorithm}
\usepackage{graphicx}
\usepackage[caption=false]{subfig}

\newcommand*{\affaddr}[1]{#1} 
\newcommand*{\affmark}[1][*]{\textsuperscript{#1}}
\newcommand*{\email}[1]{\texttt{#1}}

\begin{document}

\title{Anomaly Detection via Graphical Lasso}
\date{}
\author{ Haitao Liu\affmark[1], 
Randy C. Paffenroth \affmark[2],
Jian Zou \affmark[2],
and Chong Zhou\affmark[1] \\
\affaddr{\affmark[1] Data Science Program}\\
\affaddr{\affmark[2] Mathematical Sciences}\\
\email{\{hliu5,cpaffenroth,jzou,czhou2\}@wpi.edu}\\
\affaddr{Worcester Polytechnic Institute}}

\maketitle

\begin{abstract}
 Anomalies and outliers are common in real-world data, and they can arise from many sources, such as sensor faults. Accordingly, anomaly detection is important both for analyzing the anomalies themselves and for cleaning the data for further analysis of its ambient structure. Nonetheless, a precise definition of anomalies is important for automated detection and herein we approach such problems from the perspective of detecting sparse latent effects embedded in large collections of noisy data. Standard Graphical Lasso based techniques can identify the conditional dependency structure of a collection of random variables based on their sample covariance matrix.  However, classic Graphical Lasso is sensitive to outliers in the sample covariance matrix. In particular, several outliers in a sample covariance matrix can destroy the sparsity of its inverse. Accordingly, we propose a novel optimization problem that is similar in spirit to Robust Principal Component Analysis (RPCA) and splits the sample covariance matrix $M$ into two parts, $M=F+S$, where $F$ is the cleaned sample covariance whose inverse is sparse and computable by Graphical Lasso, and $S$ contains the outliers in $M$.  We accomplish this decomposition by adding an additional $ \ell_1$ penalty to classic Graphical Lasso, and name it ``Robust Graphical Lasso (Rglasso)''. Moreover, we propose an Alternating Direction Method of Multipliers (ADMM) solution to the optimization problem which scales to large numbers of unknowns. We evaluate our algorithm on both real and synthetic datasets, obtaining interpretable results and outperforming the standard robust Minimum Covariance Determinant (MCD) method and Robust Principal Component Analysis (RPCA) regarding both accuracy and speed.

\end{abstract}


\section{Introduction}

Gaussian Graphical Models (GGMs) are widely used to study network structures and collections of noisy data \citep{Whittaker2009, Lauritzen1996}.  Such models employ an undirected graph to represent random variables and their relationships.  If an entry $\Theta_{ij}$ of the \emph{inverse} of the covariance matrix of a collection of random variables (also called the information matrix or the precision matrix) is $0$, then variables $i$ and $j$ are conditionally linearly independent (i.e., \emph{linearly independent given all the other variables}), and this fact is represented graphically by having no edge between the nodes representing variables $i$ and $j$ \citep{Hammersley:1971aa,speed1986}.  In other words, while a covariance matrix encodes the linear predictability between pairs of random variables, the information matrix encodes the improvement in linear predictability by the addition of the given random variable, \emph{above and beyond that provided by the other random variables}.
Such graphical models are used in many application domains including causal inference, image processing, computing vision, speech recognition, analyzing gene networks, and financial analytics, to name but a few \citep{liu2017unified, zou2016efficient}.

However, when the data is high-dimensional and noisy, as is often the case in many modern applications, a direct inversion of the sample covariance matrix will often lead to a dense graph structure that masks the underlying sparse conditional dependencies \citep{meinshausen2006high, linyuan2007}.  Graphical Lasso \citep{glasso, danaher2014joint, yang2015structural, hallac2017toeplitz, hallac2017network} is the most popular model for recovering sparse information matrices from noisy data, and it proceeds by balancing a maximum likelihood principle against a sparsity-inducing $\ell_1$ regularization.

However, Graphical Lasso requires an uncorrupted sample covariance matrix as input, and even a few large anomalies (as opposed to small noise) can mask the sparsity structure of the information matrix.  Accordingly, in this paper, we provide a novel framework to identify such sparsity structures that represent latent relationships.  In particular, our work can be thought of as a generalization of current work in Robust Principal Component Analysis (RPCA) \citep{CandesRPCA, RPCAexact, Lin:2010aa, paffenroth2013space, sun2013robust} to problems of conditional dependency.  At a high level, RPCA proceeds based on the observation that a \emph{sum of a low-rank matrix and a sparse matrix} is, generally speaking, neither low-rank nor sparse.  However, under appropriate regularity conditions, one can detect whether a given matrix $M=L+S$ is a sum of a low-rank matrix $L$ and a sparse matrix $S$, and compute a unique decomposition of this form.  Similarly, the \emph{sum of a sparse matrix and a matrix whose inverse is sparse} is, generally speaking, a matrix which is neither sparse nor has a sparse inverse.  However, as we demonstrate here, one can detect whether a given matrix $M=S_1+S_2^{-1}$ is a sum of a sparse matrix $S_1$ and a matrix $F$ whose inverse is sparse $S_2 =F ^{-1}$, and compute a unique decomposition of this form. Finally, we demonstrate the interpretability of the results of applying our proposed algorithm on high-frequency financial data.

\subsection{Related Work}
Covariance estimation is very sensitive to outliers \citep{Johnstone2001} and some researchers have estimated robust covariance matrices based on the Mahalanobis distances or shape bias of a range of existing robust covariance matrix estimators \citep{rousseeuw1999fast,Filzmoser2014,hubert2014shape, Ma2001}. Some other robust covariance estimation procedures are based on cell-wise contamination data \citep{cellwise2009, ollerer2015robust, tarr2016robust} and these estimated robust covariance matrices are used as input to Graphical Lasso. Unlike these two-stage procedures, some scholars estimate the information matrix directly through a one-stage optimization with the assumption that the data follows certain distributions \citep{yang2015robust, gamma, finegold2009robust}. However, the aforementioned robust procedures for Graphical Lasso discard outliers and focus on the estimation of covariance or information matrices rather than on anomaly detection. The minimum covariance determinate (MCD) estimator is employed to construct robust Mahalanobis distances to identify local multivariate outliers \citep{Filzmoser2014}
and this is the method in the literature closest to the methods we propose. Another close method is RPCA, which decomposes the raw data matrix into a low-rank matrix $L$ representing the cleaned data and a sparse matrix $S$ containing the outliers.
In this paper, we propose a novel one-stage model to detect the outliers, which contain the hidden correlations among the variables. We demonstrate the effectiveness and efficiency of our model on three synthetic datasets, and superiority of our method over the MCD and RPCA method.  

\subsection{Contribution}

In this paper, we propose a novel algorithm that combines ideas from Graphical Lasso and RPCA to detect anomalies in large collections of noisy data.   In particular, our algorithm is similar to RPCA, in that we split the observed sample covariance matrix $M$ into two parts, $M=F+S_1$, where $F$ is the cleaned sample covariance matrix used as the input to Graphical Lasso and $S_1$ contains the outliers in $M$. However, our method differs from extent approaches in that we leverage an additional $\ell_1$ penalty applied to the input sample covariance matrix (as inspired by RPCA) in addition to the $\ell_1$ penalty classically applied by Graphical Lasso to the computed information matrix.  We call our new model the ``Robust Graphical Lasso (Rglasso)", and it has several advantages over current approaches.

\begin{itemize}

\item Our method provides a robust procedure that provides a clean sample covariance to Graphical Lasso even when the original sample covariance matrix has been contaminated by large outliers.

\item Our method can be implemented using a fast Alternating Direction Method of Multipliers (ADMM) \citep{boyd2011distributed} optimization procedure that scales to large problems with many random variables.

\item  Unlike Robust Principal Component Analysis (RPCA), we decompose the raw data matrix into a sparse matrix $S_1$ containing the outliers and a full-rank matrix $F$, whose inverse is a sparse matrix $S_2 =F^{-1}$.

\end{itemize}


\section{Background}
In this section, we provide some key ideas from Robust Principal Component Analysis that computes a unique decomposition of a matrix in the form of a sum of a low-rank matrix and a sparse matrix. We then show how this framework can be generalized to conditional dependency learning and Graphic Lasso, which is a popular model for retrieving sparse conditionally dependencies in the literature. 
\vspace{1.5mm}

\noindent \textbf{Notation} Throughout this paper, we use the following notational conventions:  $\Sigma$ is a covariance matrix,  $\Theta=\Sigma^{-1}$ is an information matrix, $\bm{X}$ are vectors, $\mathbf{V}$ and $\mathbf{E}$ are sets, and $L$, $S$ and $M$ are matrices.

\subsection{Robust Principal Component Analysis}

Robust Principal Component Analysis (RPCA) is an extension of Principal Component Analysis (PCA) that attempts to reduce the sensitivity of PCA to outliers by decomposing the raw data into outliers and the rest of the data \citep{hotelling1933analysis, eckart1936approximation, jolliffe1986principal}. After decomposition, the cleaned data can be accurately approximated by a low-rank subspace \citep{CandesRPCA}. 
In particular, the decomposition of RPCA is computed by solving the following optimization problem \citep{CandesRPCA}:
\begin{equation}
\begin{gathered}
\text{minimize}  \  \| L\|_* + \lambda \|S\|_1 \\
\text{s.t.} \  M =L + S,
\end{gathered}
\label{rpca}
\end{equation}
where the matrix $M$ is the input data, the matrix $L$ is a low-rank matrix representing the cleaned data, the matrix $S$ contains element-wise outliers, $\|\cdot\|_{*}$ is the nuclear norm (i.e., the sum of the singular values of the matrix) and $\|\cdot\|_{1}$ is the one norm (i.e., the sum of the absolute values of the entries). RPCA can detect whether a given matrix $M$ is a sum of a low-rank matrix $L$ and a sparse matrix $S$ by minimizing the nuclear norm of $L$ and the $\ell_1$ norm of $S$ \eqref{rpca}. The nuclear norm encourages the low-rankness of $L$ by summing the singular values of $L$. The $\ell_1$ norm encourages the sparsity of $S$ by summing of the absolute values of the entries. The unique decomposition of this form can be achieved without knowing the true rank of $L$ and support of $S$, assuming some mild regularity conditions \citep{CandesRPCA, Lin:2010aa, RPCAexact, paffenroth2013space}. Our description of RPCA follows the notation and outline of Zhou and Paffenroth \citep{zhou2017anomaly}.

As we will detail, our problem is similar to, but more complicated than problem \eqref{rpca}.
However, the more complicated problem can still be solved using a combination of ideas involving Alternating Direction Method of Multipliers \citep{boyd2011distributed} and other off-the-shelf solvers such as the proximal solvers for the $\ell_{1}$ penalties \citep{parikh2014proximal}.

\subsection{Graphical Lasso}
Gaussian Graphical models \citep{speed1986} estimate conditional dependency of a collection of random variables that are represented by an undirected graph $\mathbf{G} = (\mathbf{V}, \mathbf{E})$, where $\mathbf{V}$ contains $p$ vertices corresponding to $p$ random variables and $\mathbf{E}$ is an edge set $\mathbf{E} = {e_{i, j} \in \mathbf{V} |i \neq j} $ describing the \emph{conditional independence} relationship among the random variables. Any absent edge between $v_{i}$ and $v_{j}$, which indicates the variables $\bm{X}^{(i)}$ and $\bm{X}^{(j)}$ are \emph{conditionally} independent, corresponds to a zero in the \emph{inverse} of the covariance matrix of the variables (also called the information matrix) \citep{Whittaker2009, Lauritzen1996}. Such conditional independence is computed by way of the Gaussian likelihood estimation in the following optimization problem \citep{linyuan2007}:
\begin{equation}
\underset{\Theta \succ 0}{\arg\min}\ -log|\Theta|+tr(M\Theta),
\label{ggm}
\end{equation}
where $log|\cdot|$ is the log determinant, $tr(\cdot)$ denotes the trace, $M$ is the observed sample covariance matrix, and $\Theta$ is the information matrix which is
positive definite.

The standard approach to enforcing the sparsity of the information matrix is to add an $\ell_{1}$ penalty to the information matrix to form a likelihood-penalty estimation, namely Graphic Lasso \citep{glasso}, as in the following optimization problem:

\begin{equation}
\underset{\Theta \succ 0}{\arg\min}\ -log|\Theta|+tr(M\Theta)+ \rho\lVert \Theta \rVert_{1},
\label{glasso}
\end{equation}

\noindent where $\rho$ controls the sparsity of the estimated information matrix $\Theta$. When $\rho$ equals zero, the problem is equivalent to the maximum log-likelihood function \eqref{ggm}. As $\rho$ grows, more entries in the information matrix $\Theta$ are encouraged to be zero, which leads to a more sparse graph structure. When $\rho$ goes to infinity, all the off-diagonal entries in the information matrix will be zero, meaning all the variables are mutually independent of each other. \footnote{Note, the diagonal will also be zero, but that is not essential for our analysis.}

\section{Problem Definition}
\label{ProbDef}

Consider multivariate observations sampled from a distribution $x \sim N (0, \Sigma_0)$, where $\Sigma_0$ is the contaminated population covariance matrix. With such samples, we can compute the observed contaminated sample covariance matrix $M$. In some applications, $M$ is provided directly, and we don't need to compute it from the original data. Such an observed corrupted sample covariance matrix $M$ is not a good estimator of the covariance matrix of the population anymore, and cannot be processed by Graphical Lasso since Graphical Lasso is very sensitive to the outliers in the sample covariance matrix.  In particular, even a few large anomalies can mask the underlying sparsity structure of the information matrix. Here, we aim to detect the sparse latent effects embedded in the observed corrupted sample covariance $M$ rather than estimate the underlying covariance matrix $\Sigma$ and its inverse $\Theta = \Sigma^{-1}$.
Accordingly, we formulate the optimization problem to detect such hidden effects by a combination of RPCA and Graphic Lasso. Such a combination takes advantage of the anomaly detection ability of RPCA to give a clean sample covariance matrix as input to Graphic Lasso. Just as in RPCA, our method isolates the anomalies in the corrupted sample covariance matrix into a sparse matrix, and the remaining part is noise-free, whose inverse has a clear sparse structure. Specifically, we split a corrupted sample covariance matrix $M$ into two parts $M=F+S$, where $F$ represents correlations of the bulk of the data and we consider the inverse of $F$, estimated by $\Theta$, as the ``information matrix'' of the bulk of the data, of which the sparse structure is clearly captured by Graphic Lasso, and $S$ contains the anomalies in $M$. The corruption level is balanced by the $\ell_1$ norm of $S$ as compared to the loss of Graphic Lasso, as in the following optimization problem:

\begin{equation}
\begin{gathered}
\underset{\Theta \succ 0, L \succeq 0, S}{\arg\min} -log|\Theta| +tr(F\Theta) +\rho\| \Theta\|_1 + \lambda \|S\|_{1} \\
\textrm{s.t.} \ \ M = F+S.
\end{gathered}
\label{setup}
\end{equation}

\noindent In problem \eqref{setup}, $\Theta$ is the information matrix, which is positive-definite, and $M$ is the observed corrupted sample covariance matrix. $F$, which is also positive-semidefinite, is the recovered clean sample covariance matrix whose inverse is sparse and computable by Graphical Lasso. $S$ is the sparse matrix containing the hidden correlation in $M$. $\rho$ and $\lambda$ play an essential role in tuning the sparsity of the graph and the corruption level. $\rho$ controls the sparsity of the graph structure by enforcing the entries of the information matrix $\Theta$ to be zero. A choice of a larger value of $\rho$ leads to a more sparse graph structure. $\lambda$ plays a role in splitting  $M$ into $F$ and $S$ by encouraging the sparsity of $S$. The sparsity increases as the value of $\lambda$ grows. When $\lambda=0$, problem \eqref{setup} is just standard Graphical Lasso.

Note, problem $\eqref{setup}$ is not a convex optimization problem.  In particular, the term
$tr(F\Theta)$ is \emph{bi-convex}.  Accordingly, there are not current theoretical guarantees that our ADMM procedure will always return the global minimizer.  However,
as an empirical matter, we still achieve high-quality solutions (see Section \ref{num_resutls} and \ref{real_result}). 

\vspace{1.5mm}

\section{Proposed Algorithm} 

This algorithm description is inspired by Hallac et al. \citep{hallac2017network}.

To solve problem \eqref{setup} efficiently, we propose an alternating direction method of multipliers (ADMM) algorithm \citep{boyd2011distributed}. The idea of ADMM is to split a complex problem up into several subproblems. Each time, we only need to optimize a subproblem, which it is easy to handle, with all the other problems fixed. Then, we can iterate to solve each subproblem and update its solution until the user-defined converge criterion is reached. In this section, we develop analytical and closed-form solutions to the separable subproblems. These subproblem solutions are fast and easy to implement. To do so, we rewrite problem \eqref{setup} by introducing a new variable $Z$ to replace the $\Theta$ in $\ell_1$ norm so that we minimize the log-likelihood and $ \ell_1$ norm of $\Theta$ separately. The convex optimization problem becomes

\begin{equation}
\begin{gathered}
\underset{\Theta \succ 0, F \succeq 0, S, Z}{\arg\min} -log|\Theta| +tr(F\Theta) +\rho\| Z\|_1 + \lambda \|S\|_{1}  \\
\textrm{s.t.} \ \ M  = F+S \\
Z  =  \Theta.
\end{gathered}
\label{addZ}
\end{equation}

To include the two constraints into the minimization target in $\eqref{addZ}$, we write the corresponding augmented Lagrangian \citep{hestenes1969} as
\begin{equation}
\begin{gathered}
\underset{\Theta \succ 0, F \succeq 0, S}{\arg\min} -log|\Theta| +tr(F\Theta) +\rho\| Z\|_1 + \lambda \|S\|_{1} +\frac{\mu_1}{2}\| \Theta \\
- Z + U_1 \|_{F}^2 + < U_2 , M - F - S > + \frac{\mu_2}{2} \| M-F-S \|_{F}^2.
\end{gathered}
\label{aug}
\end{equation}

\noindent Writing \eqref{aug} in the forms of proximal operators \citep{parikh2014proximal} allows us to take advantage of well-known properties to find closed-form updates for each variable in the ADMM subproblems. Problem $\eqref{aug}$ is equivalent to 

\begin{equation}
\begin{gathered}
\underset{\Theta \succ 0, F \succeq 0, S}{\arg\min}  -log|\Theta| +tr(F\Theta) +\rho\| Z\|_1 + \lambda \|S\|_{1} \\
+\frac{\mu_1}{2}\| \Theta - Z + U_1 \|_{F}^2 + \frac{\mu_2}{2} \| \frac{1}{\mu_2}U_2 + M-F-S \|_{F}^2.
\end{gathered}
\label{aug1}
\end{equation}

\vspace{1.2mm}
\noindent \textbf{Proximal Operators.} For a real-valued function $f(X)$, the proximal operators is defined as

\begin{equation}
\begin{gathered}
X^{k+1} := prox_{\phi}f(X^k) \\
prox_{\phi}f(X^k) = \underset{}{\arg\min} \  f(x) +(1/2\phi)\|X-X^k\|_{F}^2,
\end{gathered}
\label{proximal}
\end{equation}

\noindent where $k$ is the iteration counter. The update value of $X$ should balance minimizing the function $f(X)$ against being close to the previous iterate $X^k$.

\vspace{1.2mm}
\noindent \textbf{ ADMM Subproblems.} As shown in Algorithm \ref{alg_rglasso}  below, our ADMM algorithm divides problem $\eqref{aug1}$ into four subproblems. Primal variables $\Theta$, $ Z $, $ F $ and $ S$ in the four subproblems are alternately optimized. After each iteration, the scaled dual variable $ U_1 $ and $ U_2 $ are also updated. The algorithm operates over the following six steps until converge criteria are satisfied. 

\begin{displaymath}
\begin{gathered}
\begin{aligned}
\Theta^{k+1} := &  \underset{\Theta \succ 0}{\arg\min}  -log|\Theta| +tr(F^k\Theta) + \frac{\mu_1}{2}\| \Theta - Z^k  + U_1^k \|_{F}^2 \\
Z^{k+1} := &  \underset{Z}{\arg\min} \  \rho\| Z\|_1 + \frac{\mu_1}{2}\| \Theta^{k+1} - Z + U_1^k \|_{F}^2 \\
U_{1}^{k+1} := & \ U_{1}^k + \Theta^{k+1} -Z^{k+1} \\
F^{k+1} := &  \underset{F \succeq 0}{\arg\min}  \ tr(F\Theta^{k+1}) + \frac{\mu_2}{2} \| \frac{1}{\mu_2}U_2^k + M-F -S^k \|_{F}^2  \\
S^{k+1} := &  \underset{S}{\arg\min} \ \lambda \|S\|_{1} + \frac{\mu_2}{2} \| \frac{1}{\mu_2}U_2^k + M-F^{k+1}-S \|_{F}^2 \\
U_{2}^{k+1} := & \ U_{2}^k +M-F^{k+1}-S^{k+1}  \\
\end{aligned}
\end{gathered}
\end{displaymath}

\noindent \textbf{Convergence Criteria.} We set two stopping criteria, $\Delta_1 = \| \Theta^{k+1} - \Theta^{k}\|_F / \|\Theta^{k} \|_F $ and $ \Delta_2 = \| M - F -S\|_F / \| M \|_F $, where criterion $\Delta_1$ measures whether the graph structure is stable, and criterion $\Delta_2$ evaluates the stability of splitting the $M$ into an $F$ and an $S$.

\vspace{1.2mm}

\noindent \textbf{Algorithm Convergence.} In the setting of two primal variables, the convergence of ADMM is guaranteed \citep{boyd2011distributed, mota2011proof}. The convergence of ADMM with more than two primal variables is an ongoing area of research \citep{mohan2014node}. Although there is no theory to guarantee the convergence of ADMM algorithm with more than two primal variables, it is often observed to converge in practice. In our case, our ADMM algorithm involves four primal variables, and we observe that our ADMM algorithm converges well (see Section \ref{num_resutls}).

\vspace{1.2mm}

\noindent \textbf{1 $\Theta$-Update}. $\Theta$-minimization has an analytical solution \citep{boyd2011distributed, Tibs2009, danaher2014joint}.

\begin{equation}
\Theta^{k+1} = \frac{1}{2\mu_1}Q (D + \sqrt{D^2 + 4 \mu_1 I})Q^T,
\label{theta_update}
\end{equation}

\noindent where $QDQ^T$ is the eigenvalue decomposition of $\mu_1(Z-U_1) - F$ \citep{boyd2011distributed}.

\vspace{1.2mm}

\noindent \textbf{2 $\bm{Z}$-Update}. We use element-wise soft thresholding \citep{parikh2014proximal, boyd2011distributed} to minimize $Z$ (see Algorithm  \ref{alg2} for the details to update $Z$).

\begin{equation}
Z^{k+1} :=   Soft_{\rho / \mu_1}(\Theta_{ij}^{k+1} + U_{1, ij}^k).
\label{z_update}
\end{equation}

\vspace{1.2mm}

\noindent \textbf{3 $\bm{F}$-Update}. Take $tr(F\Theta^{k+1}) + \frac{\mu_2}{2} \| \frac{1}{\mu_2}U_2^k + M-F-S^k \|_{F}^2 \big)$ and set the first derivative to be zero.

\begin{displaymath}
\Theta^{k+1} - \mu_2 (\frac{1}{\mu_2}U_2^k + M-F-S^k ) = 0.
\end{displaymath}

\noindent Rearrange,

\begin{displaymath}
F =  \frac{1}{\mu_2}U_2^k + M-S^k  -\frac{1}{\mu_2} \Theta^{k+1}. 
\end{displaymath}

\noindent To satisfy the constraint that $F$ is positive-semidefinite, we project $F$ to the positive-semidefinite cone. It is not the same as the idea of Wen et al. \citep{wen2010alternating}, but is similar in spirit. Specifically, we let $QDQ^T$ denote the spectral decomposition of $ \frac{1}{\mu_2}U_2^k + M-S^k  -\frac{1}{\mu_2} \Theta^{k+1}  $. $D_+$ and $D_{\_}$ are the corresponding non-negative and negative eigenvalues. 

\begin{displaymath}
QDQ^T = \begin{bmatrix}
Q_1 & Q_2 
\end{bmatrix}
\begin{bmatrix}
D_+ & 0 \\
0 & D_{\_} 
\end{bmatrix}
\begin{bmatrix}
Q_1^T \\
Q_2^T 
\end{bmatrix}
\end{displaymath}

\noindent We can get the projection of $F$ as  $F_{SPD}$

\begin{equation}
 F_{SPD} := Q_1 D_+ Q_1^T.
\end{equation}

\vspace{1.2mm}

\noindent \textbf{4 $\bm{S}$-Update}. We use element-wise soft thresholding to minimize $S$ (see Algorithm \ref{alg2} for the details to update $S$).

\begin{equation}
S^{k+1} :=   Soft_{\lambda / \mu_2}(M_{ij} - F_{ij}^{k+1} + U_{2, ij}^k).
\end{equation}

\vspace{1.2mm}

\noindent \textbf{Algorithm Initialization}\footnote{Our code is on github \url{https://github.com/lht1949/AnomalyDetection/blob/master/Rglasso.py}.}.  Since problem \eqref{setup} is not convex (see Section \ref{ProbDef}), the initialization of the internal variables of the algorithm is very important. Based on our experiments and observations, our algorithm has returned high-quality results when we initialize $F =0$  and $ S = M - F $ (see details in Algorithm \ref{alg_rglasso}).

\begin{algorithm}
	\caption{Robust Graphical Lasso}  
	\begin{flushleft}
The input of the algorithm:
\end{flushleft}
	\begin{itemize}
		\item The empirical sample covariance matrix $\mathit{M} \in \mathbb{R}^{m \times m}$.
		\item The tuning parameter $\rho$ penalizes the information matrix $\Theta$.
		\item The tuning parameter $\lambda$ penalizes the sparse latent effects matrix $\mathit{S}$.
	\end{itemize}
	
	The algorithm will use the following internal variables:
	\begin{itemize}
		\item $\mathit{S} \in \mathbb{R}^{m \times m}$, $\mathit{F} \in \mathbb{R}^{m \times m}$,  $\mathit{Z} \in \mathbb{R}^{m \times m}$, $\mathit{U}_1 \in \mathbb{R}^{m \times m}$, $\mathit{U}_2 \in \mathbb{R}^{m \times m}$, $\mathit{\mu}_1 \in \mathbb{R}$, $\mathit{\mu}_2 \in \mathbb{R}$, $\mathit{\beta } \in \mathbb{R}$, $\rho \in \mathbb{R}$, $\lambda \in \mathbb{R}$, converged $\in$ $\lbrace \text{True, False} \rbrace$
	\end{itemize}

	Initialize the variables: $\mathit{F} = 0$, $\mathit{Z}=0$, $\mathit{U_1}=0$, $\mathit{U_2}=0$, $ S = M - F $, $\mathit{\mu_1}=0.2$, $\mathit{\mu_2}=0.2$, $\mathit{\beta }=1.2$  and $converged =$ False.

\begin{flushleft}	
While (not converged):
\end{flushleft}
	\begin{enumerate}
		\item Update the values of $\mathit{\Theta}  $,  $ \mathit{Z} $, $ \mathit{F} $ and $ \mathit{S} $:
		\begin{enumerate}
			\item  $ \Theta =  $ Find\_Optimal\_$\Theta$($Z$,  $U_1$, $F$, $\mu_1$)
			\item  $ Z =   Soft_{\rho / \mu_1}(\Theta+ U_1) $
			\item  $ F =  $ Find\_Optimal\_F($M$,  $U_2$,  $ S $, $ \Theta $,$\mu_2$)
			\item  $ S =   Soft_{\lambda / \mu_2}$($M$,  $U_2$, $F$, $\lambda$, $\mu_2$) 
		\end{enumerate}
		\item Update the Lagrange multipliers, $\mathit{\mu_1}$ and $\mathit{\mu_2}$:
		\begin{description}
			\item $U_{1}= U_{1} + \Theta -Z$ 
			\item $U_{2} = U_{2} +M-F-S$  
			\item  $\mathit{\mu_1}$ =  $\mathit{\mu_1}$ * $\mathit{\beta }$
			\item  $\mathit{\mu_2}$ =  $\mathit{\mu_2}$ * $\mathit{\beta }$
		\end{description}
		\item Check for convergence:
		\begin{description}
			\item  $ \Delta_1 = \| \Theta^{k+1} - \Theta^{k}\|_F / \|\Theta^{k} \|_F $
			\item $ \Delta_2 = \| M - F -S\|_F / \| M \|_F $
			\item  If $ \Delta_1 < \epsilon$ and $ \Delta_2 <\epsilon$:
			\begin{description}
				\item $converged =$ True
			\end{description}
		\end{description}
	\end{enumerate}
\begin{flushleft}	
	$Return$  $\hat{F} = F$, $\hat{S} = S$, $\hat{\Theta} =\Theta$
\end{flushleft}	
\label{alg_rglasso}
\end{algorithm} 

\vspace{1.2mm}

\begin{algorithm}
	\caption{Soft Thresholding Operation on $Z$ and $S$}
	\begin{enumerate}
		\item  $Z^{k+1} =   Soft_{\rho / \mu_1}(\Theta_{ij}^{k+1} + U_{1, ij}^k)$
		\begin{enumerate}
			\item if $\Theta_{ij}^{k+1} + U_{1, ij}^k > \rho / \mu_1$:  \\
			$ Z^{k+1} = \Theta_{ij}^{k+1} + U_{1, ij}^k - \rho / \mu_1$
			\item if  $\Theta_{ij}^{k+1} + U_{1, ij}^k| < \rho / \mu_1$ \\
			$ Z^{k+1} = 0$
			\item if $\Theta_{ij}^{k+1} + U_{1, ij}^k < - \rho / \mu_1$ \\
			$ Z^{k+1} = \Theta_{ij}^{k+1} + U_{1, ij}^k + \rho / \mu_1$
		\end{enumerate}
		\item  $S^{k+1} =   Soft_{\lambda / \mu_2}(M_{ij} - F_{ij}^{k+1} + U_{2, ij}^k)$
		\begin{enumerate}
			\item  if $M_{ij} - F_{ij}^{k+1} + U_{2, ij}^k > \lambda / \mu_2$:  \\
			$ S^{k+1} = M_{ij} - F_{ij}^{k+1} + U_{2, ij}^k - \lambda / \mu_2 $
			\item  if $|M_{ij} - F_{ij}^{k+1} + U_{2, ij}^k| < \lambda / \mu_2$:  \\
			$ S^{k+1} = 0 $
			\item  if $M_{ij} - F_{ij}^{k+1} + U_{2, ij}^k < - \lambda / \mu_2$:  \\
			$ S^{k+1} = M_{ij} - F_{ij}^{k+1} + U_{2, ij}^k + \lambda / \mu_2 $
		\end{enumerate}
	\end{enumerate}
\label{alg2}
\end{algorithm}

\section{Numerical Study}
\label{num_resutls}

In this section, we evaluate our ADMM algorithm under three graphical structures on synthetic datasets with different anomaly setups. Further, we compare the performance of our proposed algorithm with Minimum Covariance Determinant (MCD) \citep{rousseeuw1999fast} and Robust Principal Component Analysis (RPCA) \citep{CandesRPCA} on the simulated data. 

\vspace{1.2mm}

\noindent \textbf{Graph Structure Setup.}  We test our algorithm under three network structures. The first two structures follow the setup of \citep{linyuan2007}. The third structure is intended to simulate anomalies with random placement.

\textit{Network Structure 1:} $\Theta$ has the structure that $\theta_{ii} =1$, $\theta_{i,i-1}= \theta_{i-1,i} =0.5$. \\

\textit{Network Structure 2:} $\Theta$ has the structure that $\theta_{ii} =1$, $\theta_{i,i-1}= \theta_{i-1,i} =0.5$, $\theta_{i,i-2}= \theta_{i-2,i} =0.25$. \\

\textit{Network Structure 3:} $\Theta$ has the structure that  $\theta_{ii} =1$, the positions of the off-diagonal entries are random, and 5\% of entries are non-zero.\\

\noindent \textbf{Anomalies Setup.} The true anomaly matrix $S_0$ is structured such that each row (column) has three entries. Specifically, each variable only correlates with at most two variables and at least one variable. The entries of the anomaly matrix are drawn from a normal distribution $N(\mu,10)$. We first study the case that $\mu$ is fixed at 1000, then we allow the value of $\mu$ to change.\\

\noindent \textbf{Data Generation.} The contaminated data is drawn from the multivariate distribution of $N(0, \Sigma_0)$, where $\Sigma_0=\Theta^{-1}+S_0$ is the contaminated covariance. The observed sample covariance $M$ of the contaminated data can be computed as input to our proposed algorithm.

\noindent \textbf{Experiment Result.}   In the following results, we set the number of variable $p = 200$ and the sample size $n = 100,000$.

\begin{figure} [h]
		\centering
		\includegraphics[width=0.83\textwidth]{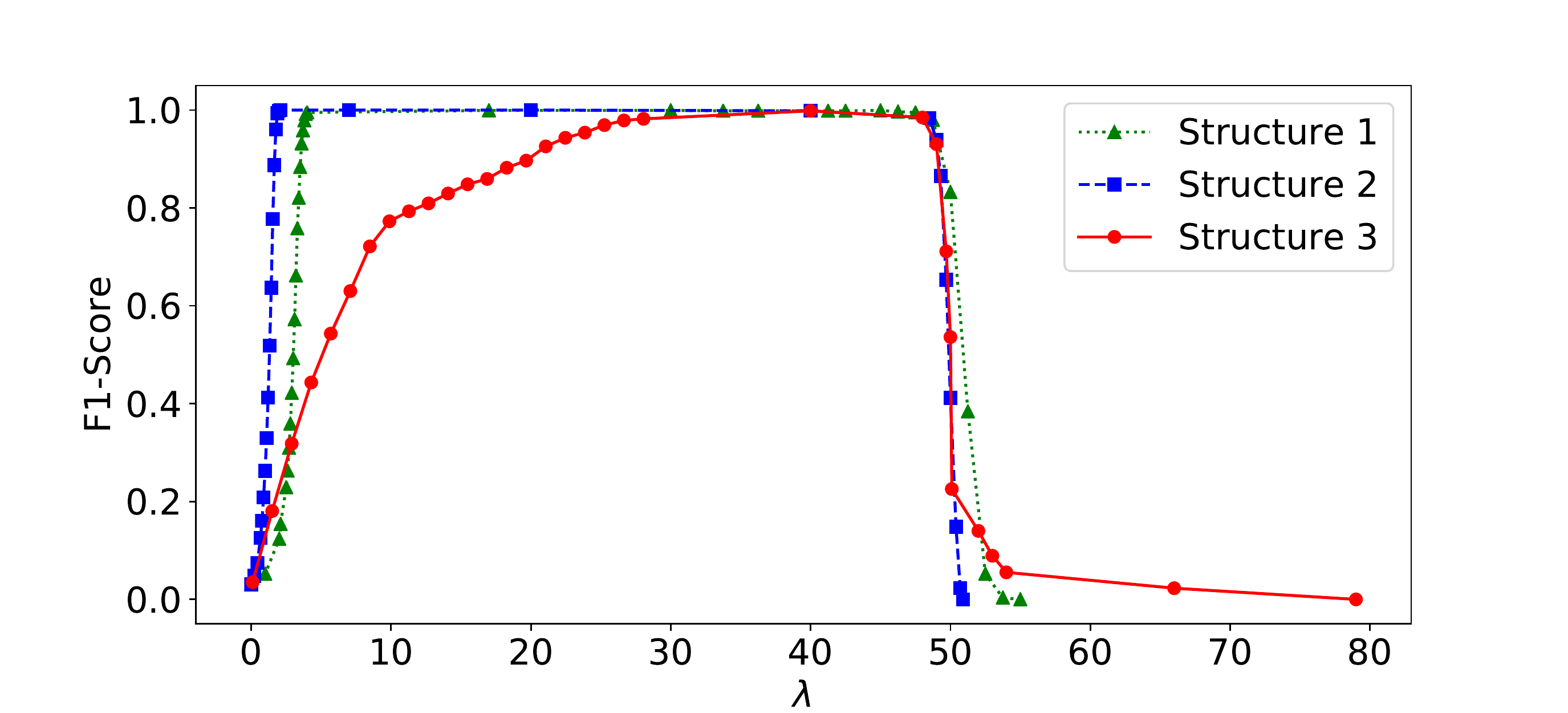}
		\caption{This figure shows that comparison of $\text{F}_1$ scores of the detected anomaly matrix $\bm{S}$ under three structures. The entries of the true anomaly matrix $\bm{S_0}$ are drawn from normal distribution $\bm{N (1000, 10)}$.}
		\label{F1_Score3structures} 
	\end{figure}

\textbf{(1) Fix $\bm{\mu = 1000}$}. We compare $\text{F}_1$ scores of the detected anomaly matrices $S$ under the three network structures (structure 1, 2, 3). Figure \ref{F1_Score3structures} indicates that the value of $\lambda$ plays a critical role in detecting the edges of the true anomaly matrix $S_0$ correctly. When the choice of $\lambda$ is small, our algorithm detects a dense anomaly matrix $S$, whose $\text{F}_1$ score is low due to overestimation of the number of edges since the true $S_0$ is sparse. As the value of $\lambda$ grows, the sparsity of the detected anomaly matrix $S$ increase correspondingly. For a certain range of values of $\lambda$, our algorithm correctly identifies the edges of the true $S_0$. As the value of $\lambda$ continues to grow, all the entries of $S$ turn to be zero, and $\text{F}_1$ score drops to zero correspondently. In addition, we can see that the curves under structures 1, 2, 3  grow at different speeds from zero to one. It indicates that our algorithm has different sensitivities to $\lambda$ at the curve increasing phase under the three structures. It is notable that all the three curves start to drop at the same value of $\lambda$. This is due to the magnitude of the anomalies dominating the magnitude of the contaminated covariance.
	
	\begin{table}[tbhp]		
		\centering
		\begin{tabular}{|c|c|c|c|c|} 
			\hline
			$\rho$	&	$\text{F}_1$ Score of S	&	$\Delta1$	&	$\Delta2$	&	Iterations	\\
			\hline
			0.001	&	0.995	&	8.97E-08	&	7.30E-14	&	50	\\
			0.005	&	0.995	&	8.11E-08	&	4.86E-14	&	51	\\
			0.01	&	0.995	&	7.92E-08	&	4.06E-14	&	52	\\
			0.05	&	0.997	&	9.17E-08	&	4.09E-14	&	53	\\
			0.1	&	0.997	&	7.06E-08	&	2.55E-14	&	54	\\
			1	&	0.997	&	6.41E-08	&	8.92E-17	&	70	\\
			2	&	0.997	&	7.50E-08	&	3.99E-17	&	75	\\
			4	&	0.997	&	8.97E-08	&	1.13E-18	&	80	\\
			\hline
		\end{tabular}
		\caption{Numerical results of structure 1 with $\bm{\lambda = 4}$. $\text{F}_1$ score of the detected anomaly matrix $\bm{S}$ is very robust to the choice of $\bm{\rho}$, which controls the sparsity of the information matrix.}
		\label{structure1}
	 \end{table}             
	
		\begin{table}[h]
		\centering
		\begin{tabular}{|c|c|c|c|c|} 
			\hline
			$\rho$	&	$\text{F}_1$ Score of S	&	$\Delta1$	&	$\Delta2$	&	Iterations	\\
			\hline
			0.001	&	0.998	&	7.15E-08	&	1.02E-14	&	49	\\
			0.005	&	0.998	&	8.73E-08	&	1.64E-14	&	49	\\
			0.01	&	0.998	&	8.73E-08	&	1.64E-14	&	49	\\
			0.05	&	0.998	&	7.79E-08	&	5.02E-15	&	50	\\
			0.1	&	0.998	&	6.57E-08	&	4.45E-15	&	53	\\
			1	&	0.998	&	9.20E-08	&	7.00E-18	&	68	\\
			2	&	0.998	&	6.76E-08	&	3.78E-20	&	74	\\
			4	&	0.998	&	8.08E-08	&	2.05E-19	&	79	\\
			\hline
		\end{tabular}
\caption{Numerical results of structure 2 with $\bm{\lambda = 1.98}$. $\text{F}_1$ score of the detected anomaly matrix $\bm{S}$ is very robust to the choice of $\bm{\rho}$, which controls the sparsity of the information matrix.}
		\label{structure2}
 \end{table}
	
	\begin{table}[h]
		\centering
		\begin{tabular}{|c|c|c|c|c|} 
			\hline
			$\rho$	&	$\text{F}_1$ Score of S	&	$\Delta1$	&	$\Delta2$	&		Iterations	\\
			\hline
			0.001	&	0.998	&	7.83E-08	&	3.76E-12	&	55	\\
			0.005	&	0.998	&	7.19E-08	&	3.69E-12	&	56	\\
			0.01	&	0.998	&	7.44E-08	&	3.66E-12	&	56	\\
			0.05	&	0.998	&	8.39E-08	&	1.07E-12	&	62	\\
			0.1	&	0.998	&	7.13E-08	&	1.07E-12	&	62	\\
			1	&	0.998	&	8.21E-08	&	2.84E-29	&	79	\\
			\hline
		\end{tabular}
		\caption{Numerical results of structure 3 with $\bm{\lambda = 36}$. $\text{F}_1$ score of the detected anomaly matrix $\bm{S}$ is very robust to the choice of $\bm{\rho}$, which controls the sparsity of the information matrix.}
		\label{structure3}
 \end{table}

    Table \ref{structure1}, \ref{structure2}, and \ref{structure3} show the numerical results of three structures with fixed values of $\lambda$. All the $\text{F}_1$ scores of the detected anomaly matrix $S$ are very robust to the choice of $\rho$, which controls the sparsity of the information matrix. In addition, our algorithm converges with a very high accuracy under all three structures. The convergence criterion $\Delta2$ is much smaller than convergence criterion $\Delta1$. In addition, the number of iterations the algorithm takes to satisfy the two convergence criteria is less than 100. As the choice of $\rho$ grows, the number of iterations grows slowly.

\begin{figure} [H]
		\centering
		\includegraphics[width=0.83\textwidth]{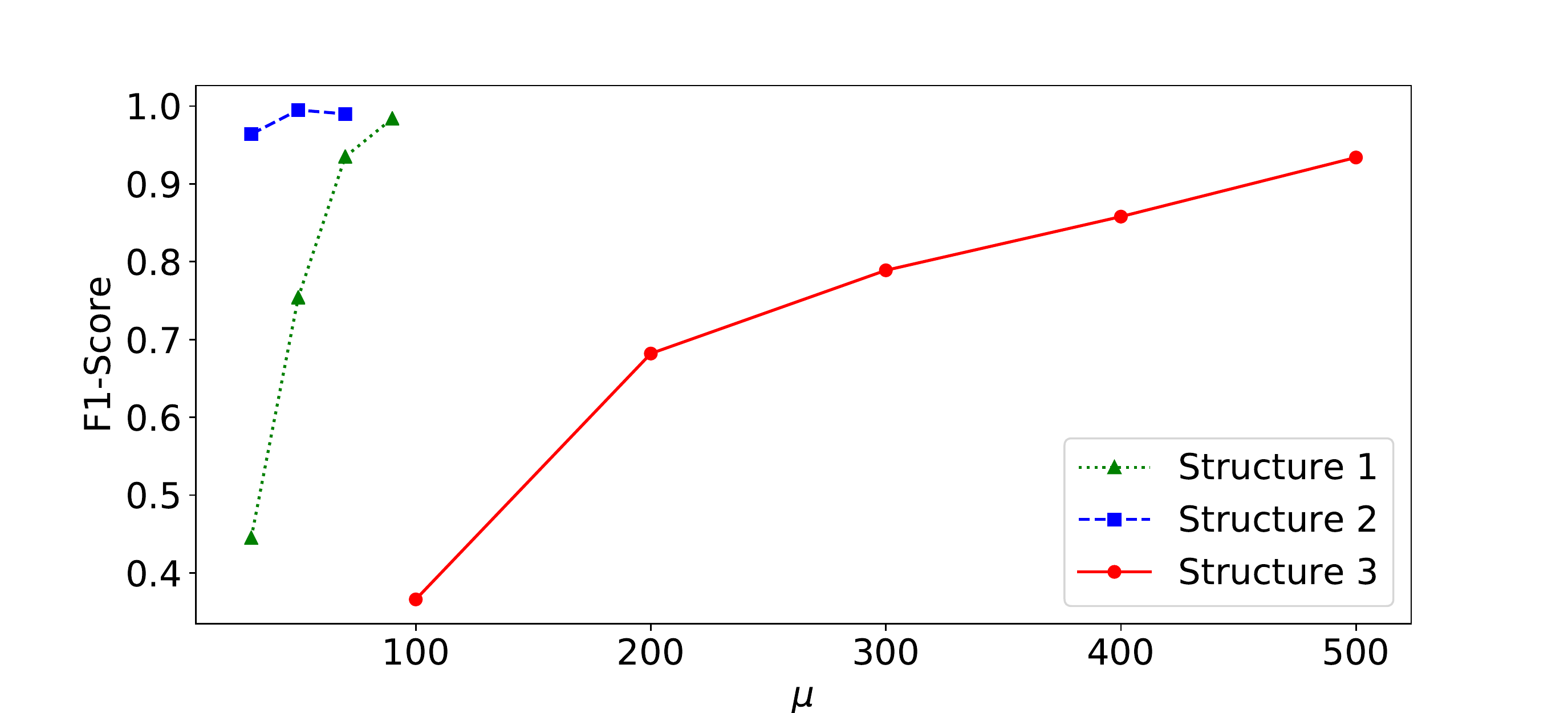}
		\caption{This figure shows that comparison of $\text{F}_1$ score of anomaly matrix $S$ under three structures (The information matrices of structure 1 and 2 have tridiagonal and five-diagonal structure. The information matrices of structure 3 is random with 5\% non-zero entries). The entries of the true anomaly matrix $S_0$ are drawn from normal distribution $N (\mu, 10)$, and $\mu$ varies.}
		\label{different_mu} 
	\end{figure}

 \textbf{(2) Vary $\bm{\mu}$}. In this part, we allow the magnitude, $\mu$, of the true anomalies to change. The $\text{F}_1$ score curves of our detected anomaly matrix under the different magnitude of the true anomalies are presented in Figure \ref{different_mu}. In general, as the value of  $\mu$ increases, our algorithm can detect the anomalies more accurately in all the structure settings. Specifically, our algorithm can separate the anomalies of much smaller magnitude from the observed corrupted sample covariance correctly under the structure 1 and 2 than structure 3 setting. 
	
\begin{figure}[H]
    \centering
    \begin{tabular}{ccc}
    \subfloat[Distribution under structure 1 with $\bm{\mu = 50}$.]{\label{structure1dist} \includegraphics[width=.3\linewidth]{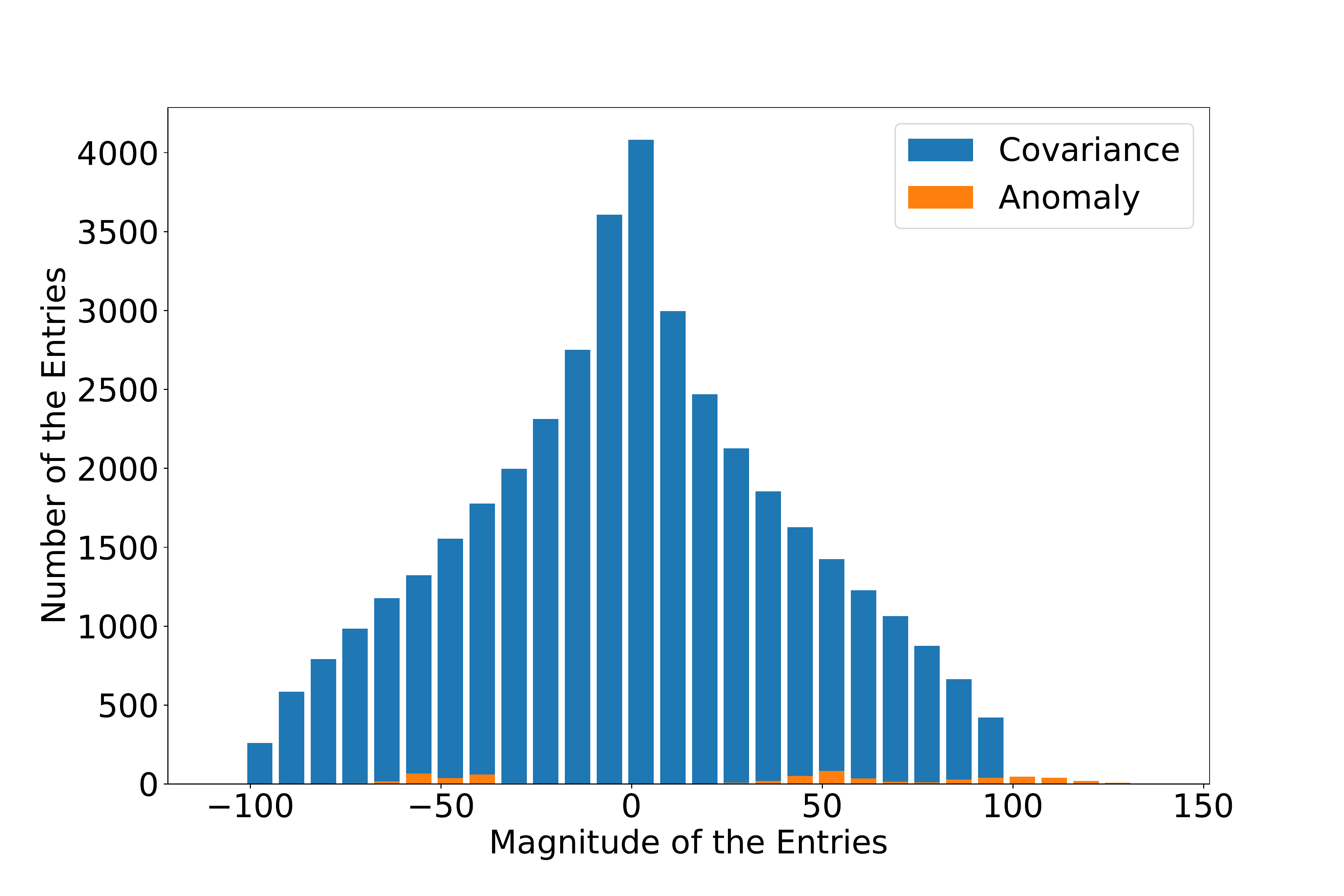}}	 &
	\subfloat[Distribution under structure 2 with $\bm{\mu = 30}$.]{\label{structure2dist} \includegraphics[width=.3\linewidth]{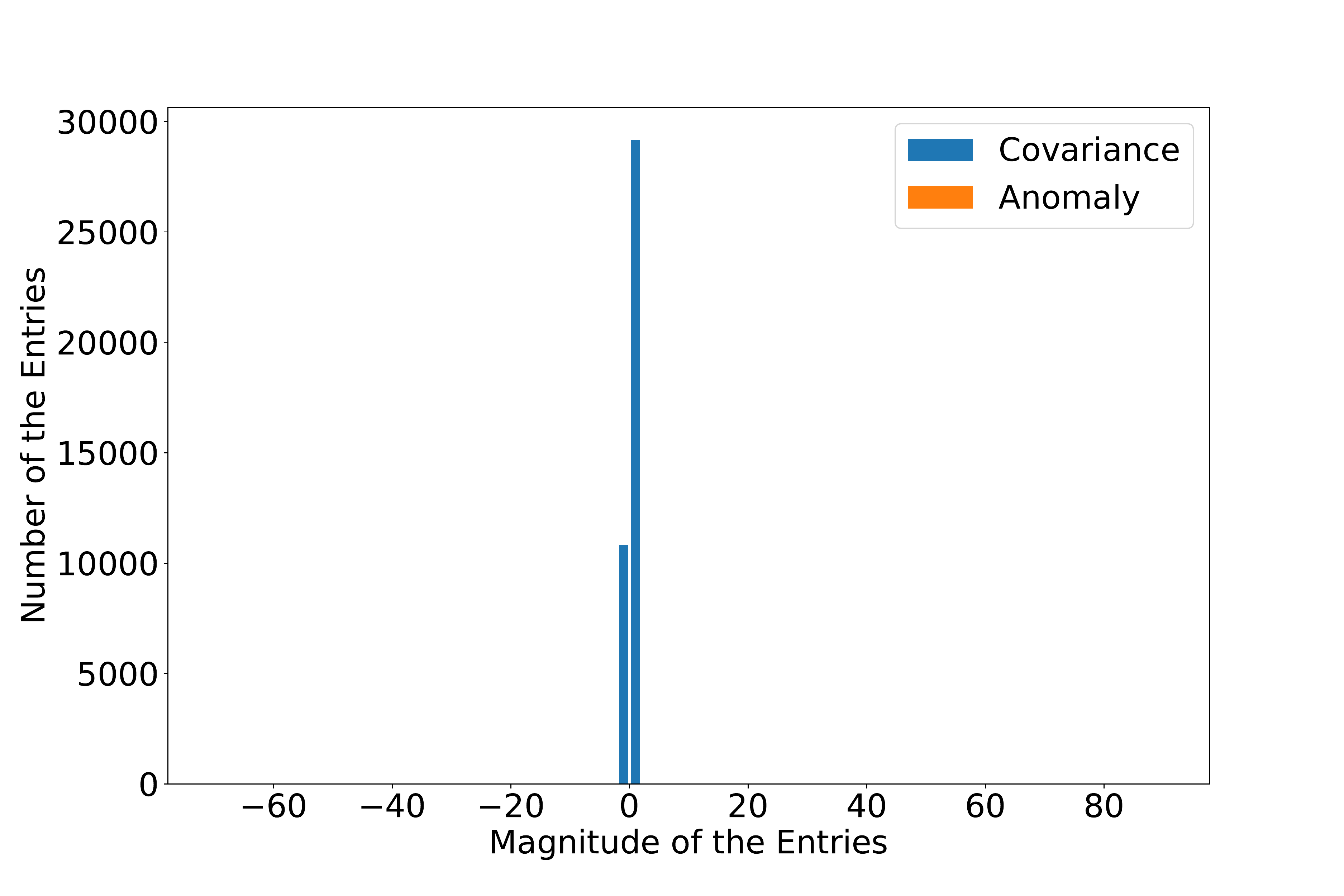}} &
		\subfloat[Distribution under structure 3 with $\bm{\mu = 200}$.]{\label{structure3dist} \includegraphics[width=.3\linewidth]{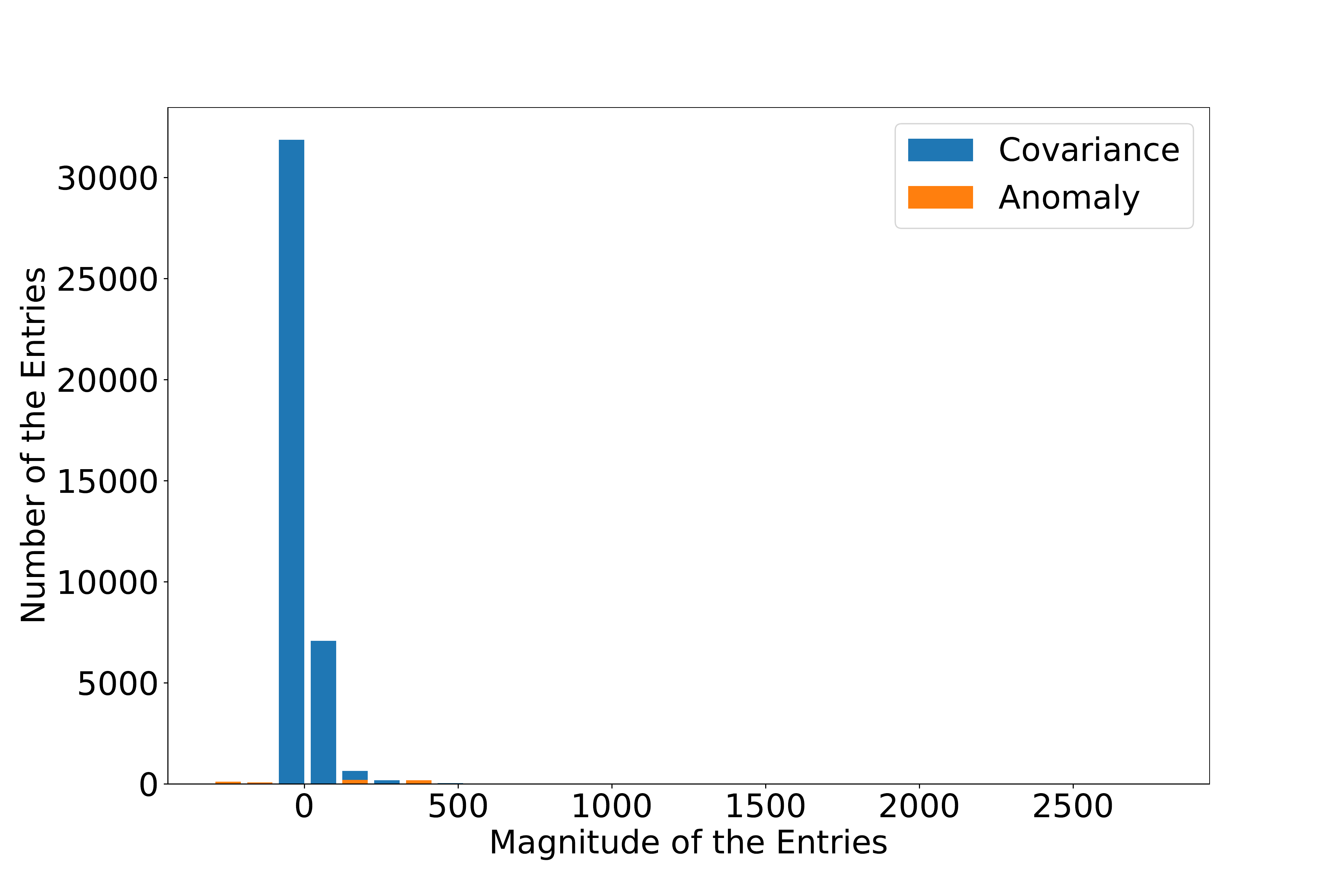}}\\
			\subfloat[Magnified distribution under structure 1.]{\label{structure1distzoom} \includegraphics[width=.3\linewidth]{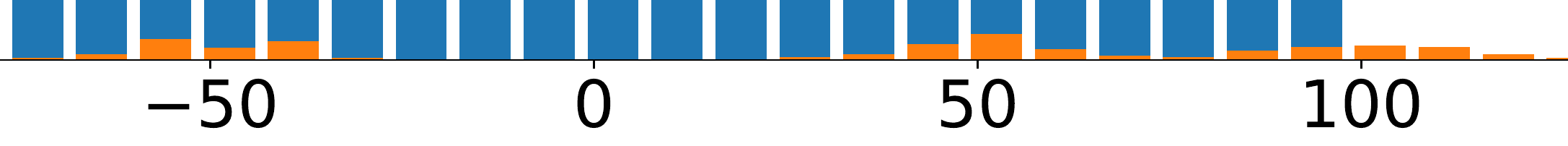}} &
			\subfloat[Magnified distribution under structure 2.]{\label{structure2distzoom} \includegraphics[width=.3\linewidth]{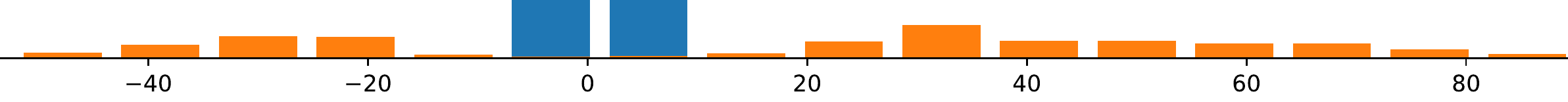}} &
		\subfloat[Magnified distribution under structure 3.]{\label{structure3distzoom}  \includegraphics[width=.3\linewidth]{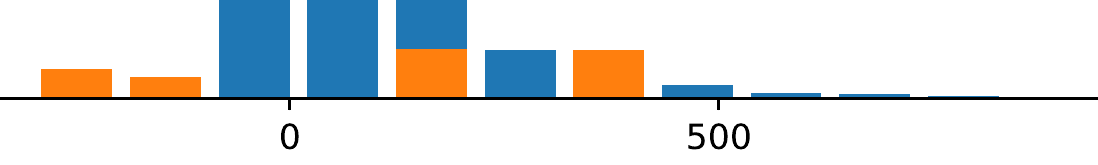}}
		\end{tabular}
		\caption{The distributions of the entries of the true covariance $\bm{\Sigma}$ and the true anomaly matrix $\bm{S_0}$ are labeled by blue and orange color correspondingly, and the entries of the anomaly matrix are drawn from $\bm{N(\mu, 10)}$. The performances of our algorithm are considered well under structure 1 (\protect\subref{structure1dist}, \protect\subref{structure1distzoom}) and 3 (\protect\subref{structure3dist}, \protect\subref{structure3distzoom}). The performance under structure 2 (\protect\subref{structure2dist}, \protect\subref{structure2distzoom}) is not considered as well as structure 1 and 3.}
		\label{Distribution}
	\end{figure}

Next, to investigate the identifiability of the true covariance matrix $\Sigma$ and the true anomaly matrix $S_0$, we examine the distributions of the entries of the true covariance matrix and the true anomaly matrix under the three structure settings. Figure \ref{Distribution}\subref{structure1dist}, \ref{Distribution}\subref{structure2dist} and \ref{Distribution}\subref{structure3dist} show that the distributions of the entries of the true covariance matrix on the top (labeled by blue color) and the true anomaly matrix at the bottom (labeled by orange color). Since the anomaly matrix is sparse with only 600 non-zero entries (out of total 40,000 entries), and the covariance matrix is dense with around 40,000 non-zero entries, the distribution is dominated by the values of the entries of the covariance matrix, and the distribution of the values of entries of the anomaly matrix is hidden. The magnified distributions in Figure \ref{Distribution}\subref{structure1distzoom}, \ref{Distribution}\subref{structure2distzoom} and \ref{Distribution}\subref{structure3distzoom} give clearer pictures of the distributions of the values of the entries of the anomaly matrix. We set $\mu = 50, 30$ and $200$ for structures 1, 2 and 3 correspondingly. Under such $\mu$ settings, we can identify the anomalies with an $\text{F}_1$ score around 0.7. The distribution of the entries of the covariance matrix under structure 1 is bell-shaped, and there are around three thousand entries with values close to or larger than that of the anomaly matrix. This suggests that our anomaly detection algorithm performs very well when the magnitude of the anomalies is smaller than that of the truth covariance under the structure 1 setup. Figure \ref{Distribution}\subref{structure2dist} shows that all the values of entries of the covariance matrix are very close to zero under the mode 2 setup. If we magnify the distribution of the entries of the anomaly matrix, Figure \ref{Distribution}\subref{structure2distzoom} indicates that our algorithm performs well under structure 2 with the condition that the magnitude of the anomalies is much larger than that of the true covariance. Under the structure 3 settings, the distribution of the values of the entries of the true covariance is a long thin tail with majority values close to zero. The magnified distribution in Figure \ref{Distribution}\subref{structure3distzoom} shows that the magnitude of the anomalies is comparable to the magnitude of the true covariance matrix.

\begin{figure}[H]
			\centering
			\subfloat[Accuracy comparison over different $p$ with fixed $n$.]{\label{F1_Score_fixN} \includegraphics[width=.48\linewidth]{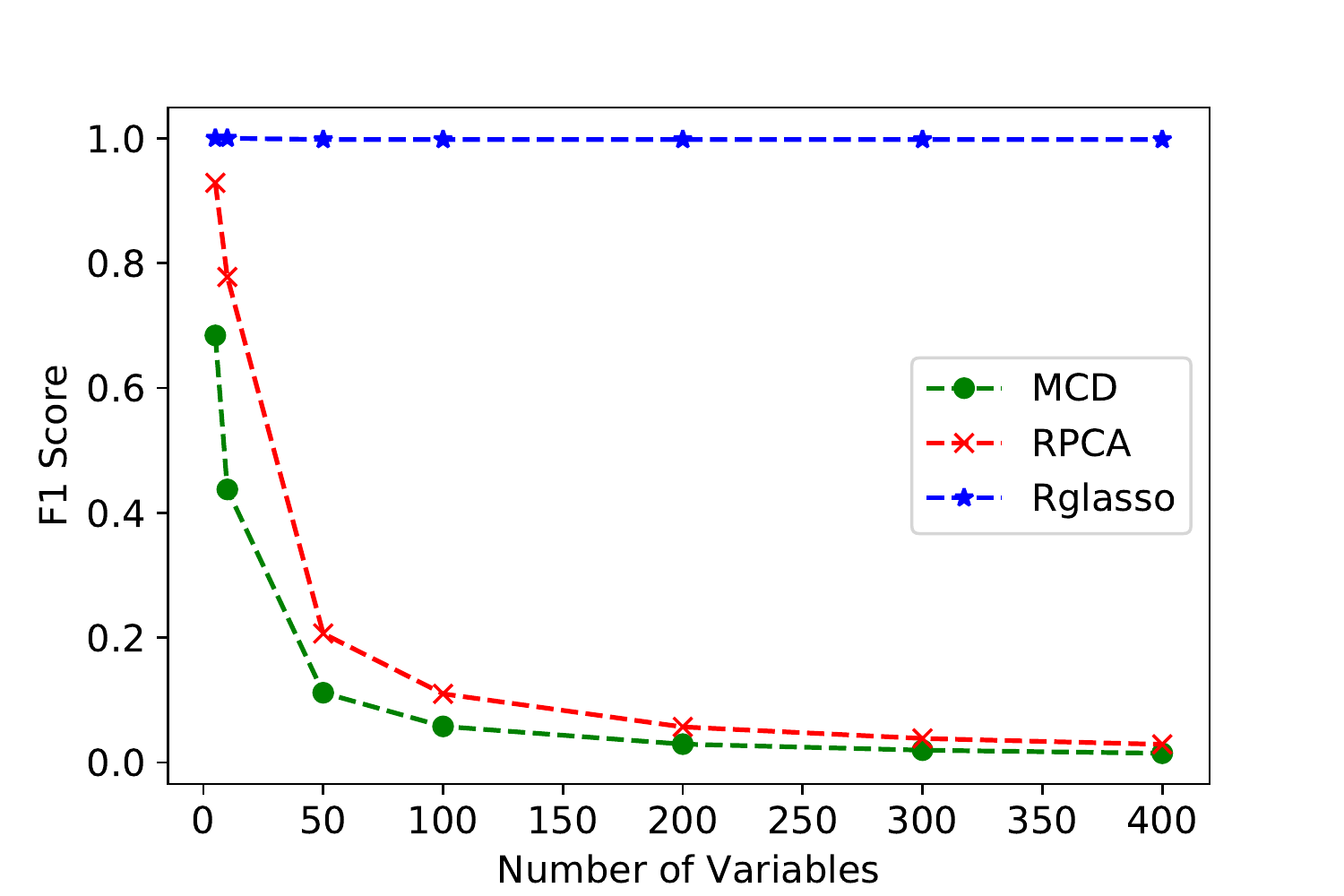}}
		\subfloat[Time cost comparison over different $p$ with fixed $n$.]{\label{time_fixN} \includegraphics[width=.48\linewidth]{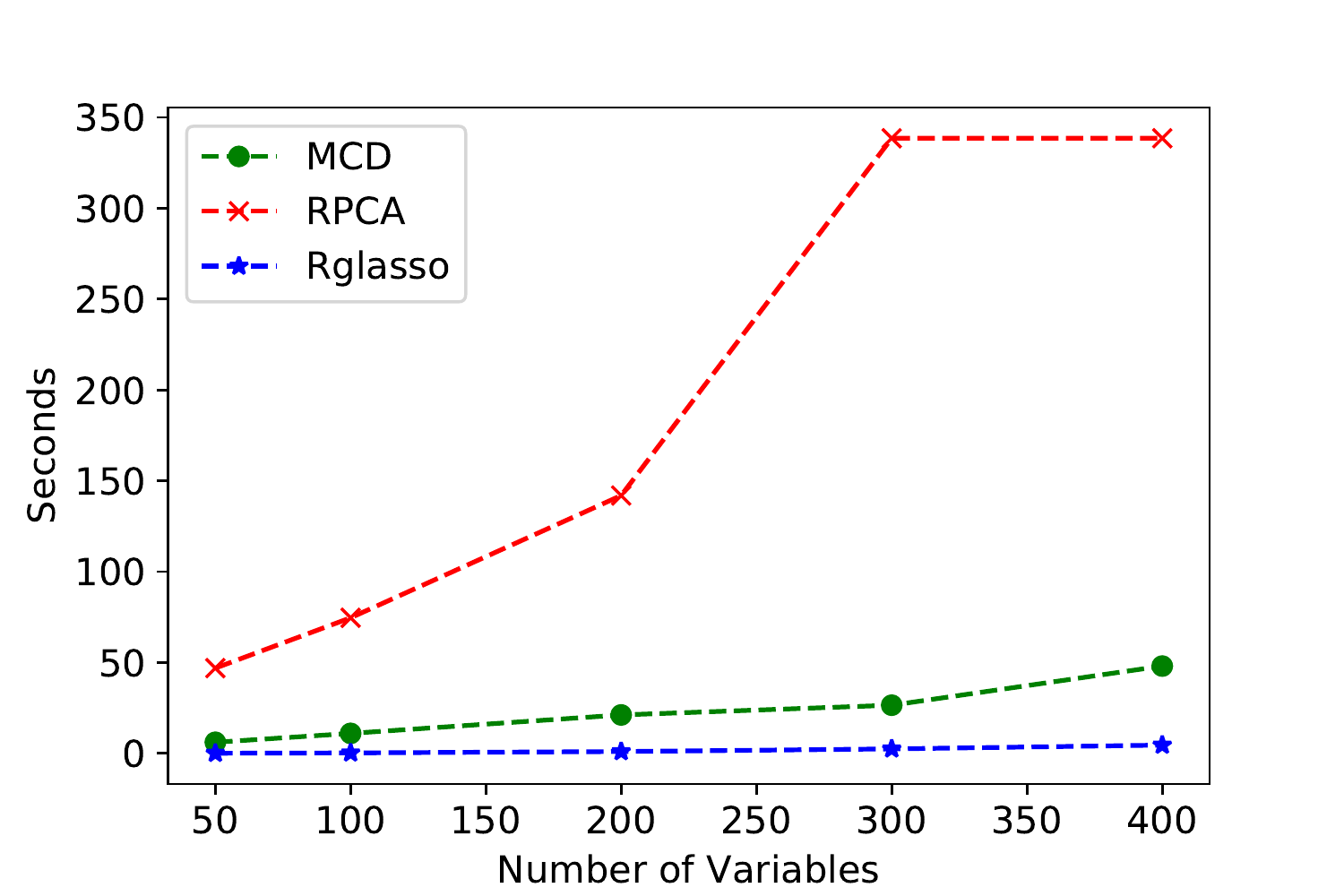}}
				\caption{Accuracy and time cost comparison of our proposed method (Rglasso) v.s. MCD method over different numbers of features ($\bm{p}$) with fixed sample size ($\bm{n=10,000}$) (\protect\subref{F1_Score_fixN} and \protect\subref{time_fixN}). The comparison is under the structure 1 setting with $\bm{\mu = 1000}$. The comparison is done in Python 2.7.14 on the MacBook Pro (Mid 2015) with processor 2.2 GHz Intel Core i7, Memory 16 GB 1600 MHz DDR3, and macOS High Sierra (Version 10.13.3). We use the MCD implementation in the package of scikit-learn (Version 0.19.0), and implement RPCA according to the algorithm of \citep{CandesRPCA}.}
	\label{comp}
	\end{figure}

\noindent \textbf{Time Cost and Accuracy.} We compare our proposed algorithm with MCD and RPCA method under the structure 1 setup with $\mu =1000$. The comparison is done in Python 2.7.14 on the MacBook Pro (Mid 2015) with processor 2.2 GHz Intel Core i7, Memory 16 GB 1600 MHz DDR3, and macOS High Sierra (Version 10.13.3). We use the MCD implementation in the package of scikit-learn 0.19.0. and implement RPCA according to the algorithm of  \citep{CandesRPCA}. Figure \ref{comp}\subref{F1_Score_fixN} and \ref{comp}\subref{time_fixN} show the results over different numbers of features ($p$) with fixed sample size ($n=10,000$). In the case that $p$ is small, MCD and RPCA can detect the anomalies with relatively high accuracy. As the value of $p$ grows, the accuracy of MCD and RPCA drop quickly to close to zero. RPCA performs poorly since it decomposes the input matrix into a low-rank matrix and a sparse matrix, however, the true input matrix is composed of a full-rank matrix and a sparse matrix. In contrast, the performance of our algorithm is very stable with $\text{F}_1$ score close to one. As to computational cost, our algorithm is much faster than MCD and RPCA as the value of $p$ grows (see Figure \ref{comp}\subref{time_fixN}). RPCA converges slowly due to the wrong decomposition of the imput matrix.
\section{Case study with application in financial data}
\label{real_result}

\begin{figure} [tbhp]
		\centering
		\includegraphics[width=0.75\textwidth]{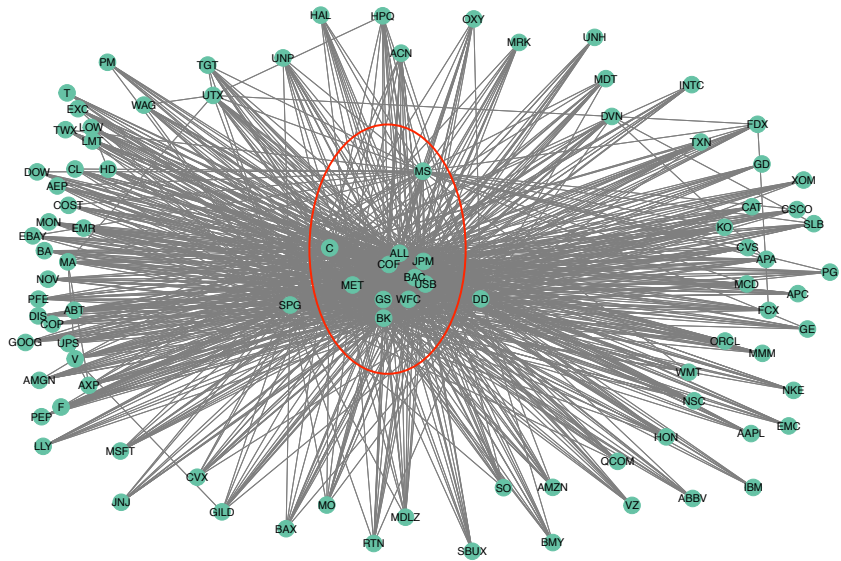}
		\caption{Latent network structure our algorithm identifies using 94 stocks in S$\&$P 100.}
		\label{All_sp100} 
\end{figure}

\begin{figure} [H]
		\centering
		\includegraphics[width=0.7\textwidth]{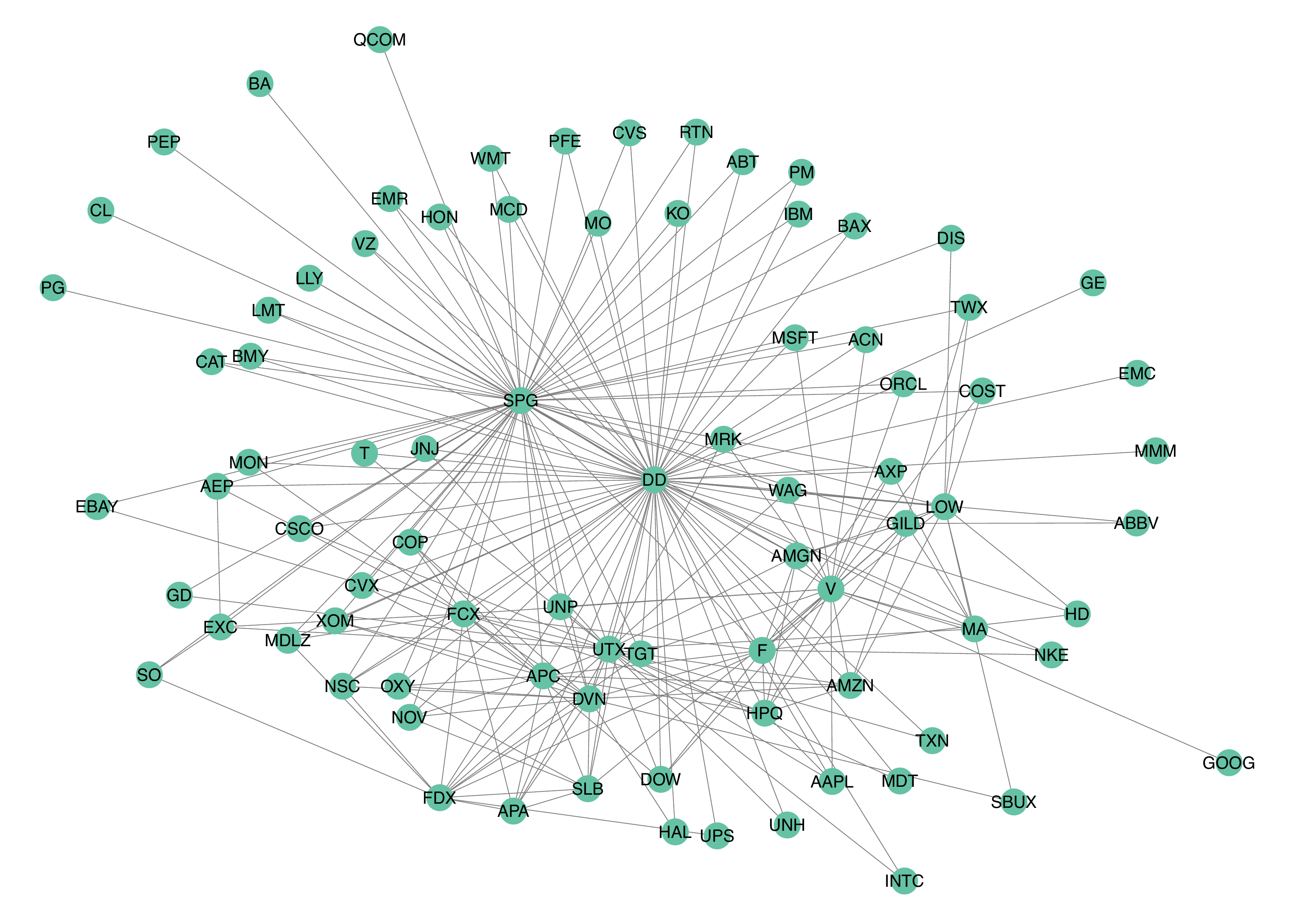}
		\caption{Latent network structure our algorithm identifies using 83 stocks (excluding the stocks in finance industry) in S$\&$P 100.}
		\label{sp100_no_finance} 
\end{figure}

\begin{figure} [tbhp]
		\centering
		\includegraphics[width=0.5\textwidth]{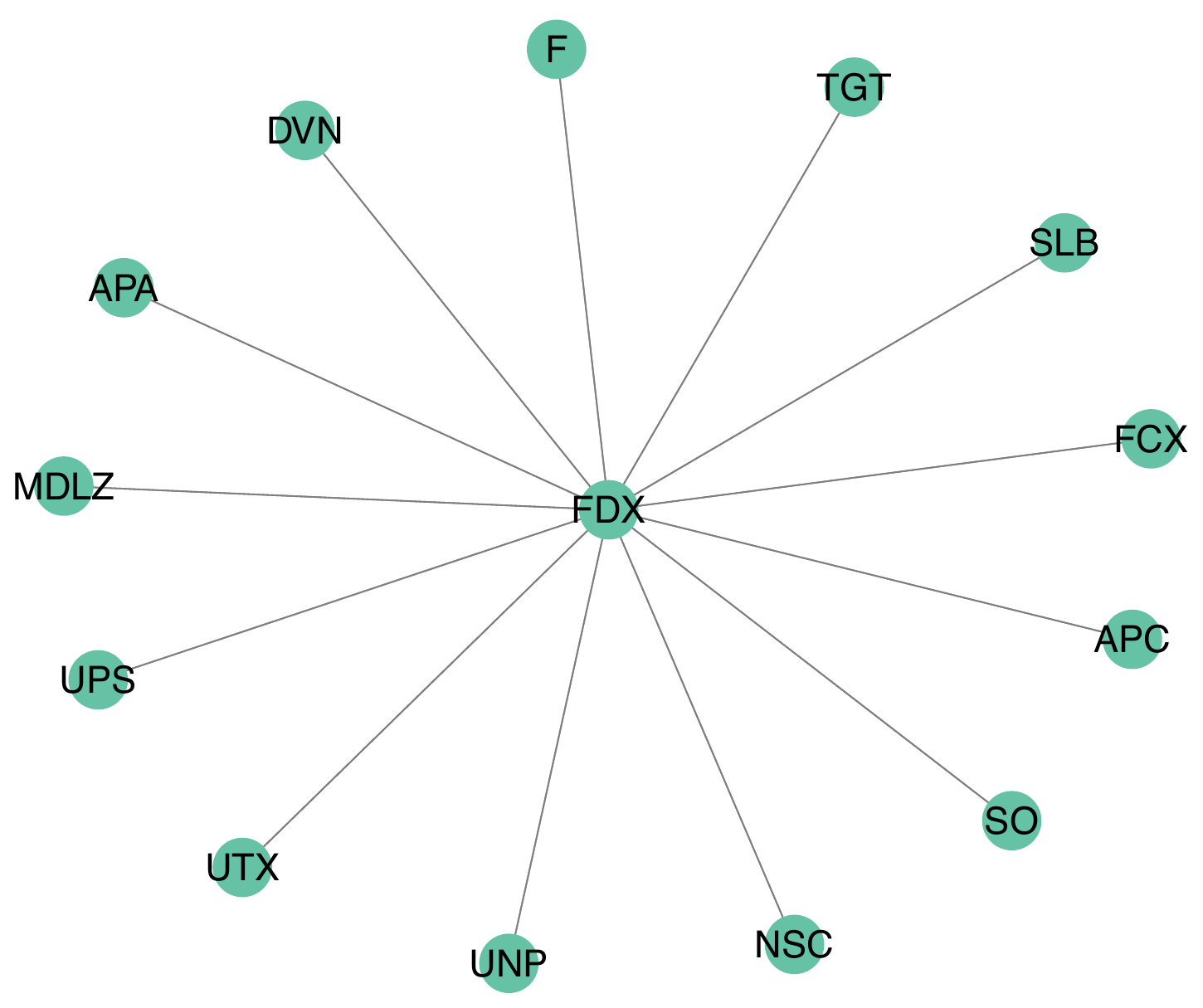}
		\caption{Latent network structure of FDX (FedEx), which is extracted from Figure \ref{sp100_no_finance}.}
		\label{FDX_example} 
\end{figure}

Detecting anomalies in stock prices is an interesting application of our methods since this usually implies the hidden connections between different companies. Such a sparse hidden graph structure reveals the latent correlation of stocks, and it can help improve portfolio allocation and better hedge risks by avoiding stocks with such hidden connections. We can infer hidden graph structures \footnote{The hidden graph structure is detected anomaly matrix $S$. Figure \ref{All_sp100}, \ref{sp100_no_finance}, and \ref{FDX_example} are visualizations by treating $S$ as adjacent matrices.} by applying our algorithm to the stock prices through their log returns. We study the intraday high-frequency transactions \footnote{The stock prices of the transactions are download from Wharton Research Data Services (WRDS).} (millisecond) of 94 stocks \footnote{We use the data in Liu et al. \citep{bicluster2018}} in S$\&$P 100 \citep{bicluster2018}. Specifically, we use 1-minute averaged log returns of the stock prices from these (millisecond) transactions on August 21, 2013. The dimension of the data matrix is 389 by 94. We compute the sample covariance of the 94 stocks, which is the input to our algorithm.  

Figure \ref{All_sp100} shows the latent network structure our algorithm identifies. We can see that the stocks with the most connections are in the center of the graph. In particular, the stocks in the red circle are all in the finance industry. These finance companies provide loans, credit lines or other financial services to other companies. Therefore, they have the most edges connected with other companies. Such wide connections result in a relatively dense graph structure. 

We further examine the data by excluding the finance stocks (MS, C, ALL, JPM, COF, BAC, USB, MET, GS, WFC and BK) in the red circle. The sparsity of the hidden graph structure we detect improves in Figure \ref{sp100_no_finance}. SPG (Simon Property Group) and DD (DuPont) have the most edges and locate in the center since SPG (Simon Property Group), as a real estate company, is correlated with other companies via office building renting, warehouse cost and etc., and DD (DuPont) as a conglomerate involves many industries.  

Figure \ref{FDX_example} shows the hidden structure of FDX (FedEx), which is extracted from Figure \ref{sp100_no_finance}. FDX (FedEx) has hidden connections with companies in 3 major industries: energy industry such as SLB (Schlumberger Business Consulting), FCX (Freeport-McMoRan Inc.), APC (Anadarko Petroleum), APA (Apache Corporation) and DVN (Devon Energy Corp), since energy consumption is the biggest cost of FDX (FedEx) as a transportation company;  transportation industry such as SO (Southern Company), NSC (Norfolk Southern), UNP (Union Pacific Corporation) and UPS (United Parcel Service), since these companies are FDX (FedEx)'s competitors; automaker/aircraft maker industry such as F (Ford Motor) and UTX (United Technologies), since FDX (FedEx) may need to purchase trucks or airplanes from automaker/aircraft makers as transportation tools.

\section{Conclusion and Future Work}
	
In this paper, we propose a Robust Graphical Lasso (Rglasso) to detect sparse latent effects via Graphical Lasso. The algorithm is similar in spirit to Robust Principal Component Analysis (RPCA). Moreover, we provide an Alternating Direction Method of Multipliers (ADMM) solution to the optimization problem which scales to large problems with many random variables. We evaluate the proposed algorithm on both real and synthetic datasets, obtaining interpretable results and outperforming the standard robust minimum covariance determinant (MCD) and RPCA method in terms of both accuracy and speed. In future work, we will explore more graphical structures and anomaly settings to understand our algorithm working conditions. In addition, we will modify the loss function and constraints to estimate the information matrix. Moreover, since we have demonstrated that our algorithm provides high-quality results both in synthetic and real data, we will explore the theoretical foundation for the convergence of the algorithm to a global minimizer.

\bibliographystyle{abbrvnat}
\bibliography{bibliography.bib}

\end{document}